\documentclass[lettersize,journal]{IEEEtran}
\usepackage{amsmath,amsfonts}
\usepackage{algorithmic}
\usepackage{algorithm}
\usepackage{array}
\usepackage[caption=false,font=normalsize,labelfont=sf,textfont=sf]{subfig}
\usepackage{textcomp}
\usepackage{stfloats}
\usepackage{url}
\usepackage{verbatim}
\usepackage{graphicx}
\usepackage{cite}
\usepackage{microtype}
\usepackage{graphicx}
\usepackage{booktabs} 
\usepackage{multirow}

\usepackage[colorlinks,linkcolor=black, filecolor=black, urlcolor=black, citecolor=black]{hyperref}
\usepackage[T1]{fontenc}

\usepackage{xkeyval}  
\usepackage{wrapfig}  
\usepackage{xparse}

\makeatletter
\newwrite\authorbibfile
\AtBeginDocument{%
  \immediate\openout\authorbibfile=\jobname.aub%
}
\AtEndDocument{%
  \immediate\closeout\authorbibfile
  \InputIfFileExists{\jobname.aub}{}{}
}

\define@key{authorbib}{scale}[1]{\def\AuthorbibKVMacroScale{#1}}
\define@key{authorbib}{wraplines}[10]{\def\AuthorbibKVMacroWraplines{#1}}
\define@key{authorbib}{imagewidth}[4cm]{\def\AuthorbibKVMacroImagewidth{#1}}
\define@key{authorbib}{overhang}[10pt]{\def\AuthorbibKVMacroOverhang{#1}}
\define@key{authorbib}{imagepos}[l]{\def\AuthorbibKVMacroImagepos{#1}}
\makeatother

\presetkeys{authorbib}{imagepos=l, imagewidth=4cm, wraplines=8, overhang=20pt}{}
\newlength{\AuthorbibTopSkip}
\newlength{\AuthorbibBottomSkip}
\NewDocumentCommand{\authorbibliography}{+o+m+m+m}{%
  \IfNoValueTF{#1}{}{%
    \setkeys{authorbib}{#1}%
    \immediate\write\authorbibfile{%
      \string\begin{wrapfigure}[\AuthorbibKVMacroWraplines]{\AuthorbibKVMacroImagepos}[\AuthorbibKVMacroOverhang]{\AuthorbibKVMacroImagewidth}^^J
        \string\includegraphics[scale=\AuthorbibKVMacroScale]{#2}^^J
        \string\end{wrapfigure}^^J
    }%
  }%
  \IfNoValueTF{#3}{%
    \typeout{Warning: No author name}%
  }{%
      \immediate\write\authorbibfile{%
      \unexpanded{\vspace{\AuthorbibTopSkip}}^^J
      \string\noindent\relax
      \unexpanded{\textbf{#3}\par}^^J
      \string\noindent\relax
      \unexpanded{#4}^^J%
      \unexpanded{\vspace{\AuthorbibBottomSkip}}^^J
      }%
  }%
}

\begin{document}

\title{Aerial-R1: Reinforcing Aerial Reasoning and Segmentation in MLLMs}

\author{Jisheng Dang, Huaiyuan Peng, Yixuan Wang, Kun Wang, Zhang Li, Tat-Seng Chua
\thanks{This work was supported in part by the National Natural Science Foundation of China (62227807), in part by the National Natural Science Foundation of China(Grant No.U24B20186), and in part by Scientific and Technological Innovation 2030 Project of MOST, China (No. 2021ZD0200600, No.2021ZD0200408), and supported by the Super-computing Center of Lanzhou University.}
\thanks{Jisheng Dang, Huaiyuan Peng, and Yixuan Wang are with School of Information Science \& Engineering, Lanzhou University, Lanzhou 730000, China.  (E-mail: dangjsh@mail2.sysu.edu.cn, penghy2024@lzu.edu.cn).}
\thanks{Kun Wang and Zhang Li are with the School of Aeronautical and Astronautical Science
and Technology, National University of Defense
Technology, Changsha 410073, China (E-mail: wangkun21@nudt.edu.cn, lizhang08@nudt.edu.cn).}
  
\thanks{Tat-Seng Chua is with the School of Computing, National University of Singapore, Singapore 119077 (E-mail: dcscts@nus.edu.sg). }
\thanks{*Corresponding author: Zhang Li.}}

\maketitle

\begin{abstract}
While Multimodal Large Language Models (MLLMs) have revolutionized general visual perception, existing methods suffer from severe hallucination in the aerial domain, where extreme scale variations and visual ambiguity abound. 
Existing Supervised Fine-Tuning (SFT) paradigms force models to mimic ground truth without modeling uncertainty, often resulting in blind guessing on tiny or unrecognizable objects in aerial scenes.
In this paper, we present Aerial-R1, the first aerial MLLM framework to reinforce visual reasoning capabilities for aerial reasoning and segmentation. 
Drawing inspiration from the reasoning-incentivizing mechanisms of DeepSeek-R1, we find that trustworthy perception stems from consistency across views, establishing a visual parallel to the Chain-of-Thought in logic.
To effectively adapt policy optimization for non-autoregressive mask generation, we introduce Hybrid-View Group Advantage-Weighted Regression (H-GAWR). This strategy constructs dynamic groups of high-resolution and low-resolution views, seamlessly bridging group-relative advantage estimation with advantage-weighted policy extraction.
By enforcing consensus on clear inputs and strictly penalizing hallucinations on ambiguous ones, Aerial-R1 develops an explicitly incentivized refusal mechanism. 
Extensive experiments on VSAI-Ref, the first aerial MLLM benchmark dedicated to jointly evaluating dense reasoning and hallucination robustness, demonstrate that Aerial-R1 significantly outperforms SFT baselines in segmentation precision while substantially mitigating fatal hallucinations in low-information scenarios, paving the way for more reliable and trustworthy aerial agents. Codes are available on \href{https://github.com/phy709/Aerial-R1}{https://github.com/phy709/Aerial-R1}.
\end{abstract}


\section{Introduction}
\label{sec:intro}

The rapid evolution of Multimodal Large Language Models (MLLMs) has revolutionized visual grounding and understanding tasks. Pioneering works such as LLaVA \cite{liu2023llava}, MiniGPT-4 \cite{zhu2023minigpt}, and Qwen-VL \cite{bai2023qwenvl} have demonstrated remarkable capabilities in aligning visual encoders with powerful Large Language Models (LLMs). Building on this, referring segmentation models like LISA \cite{lai2024lisa}, Shikra \cite{chen2023shikra}, and Sa2VA \cite{yuan2025sa2va} enable users to segment objects using natural language queries. However, applying these general-purpose models to the aerial domain, often termed the God's Eye View, remains a formidable challenge. Unlike natural scenes, aerial imagery is characterized by extreme scale variations, densely packed tiny objects, and complex background clutter, as highlighted in foundational datasets like DOTA \cite{ding2018dota}, iSAID \cite{waqas2019isaid}, and VisDrone \cite{zhu2019visdrone}.
\begin{figure*}[t]
\vskip 0in
\begin{center}
\centerline{\includegraphics[width=1.05\textwidth]{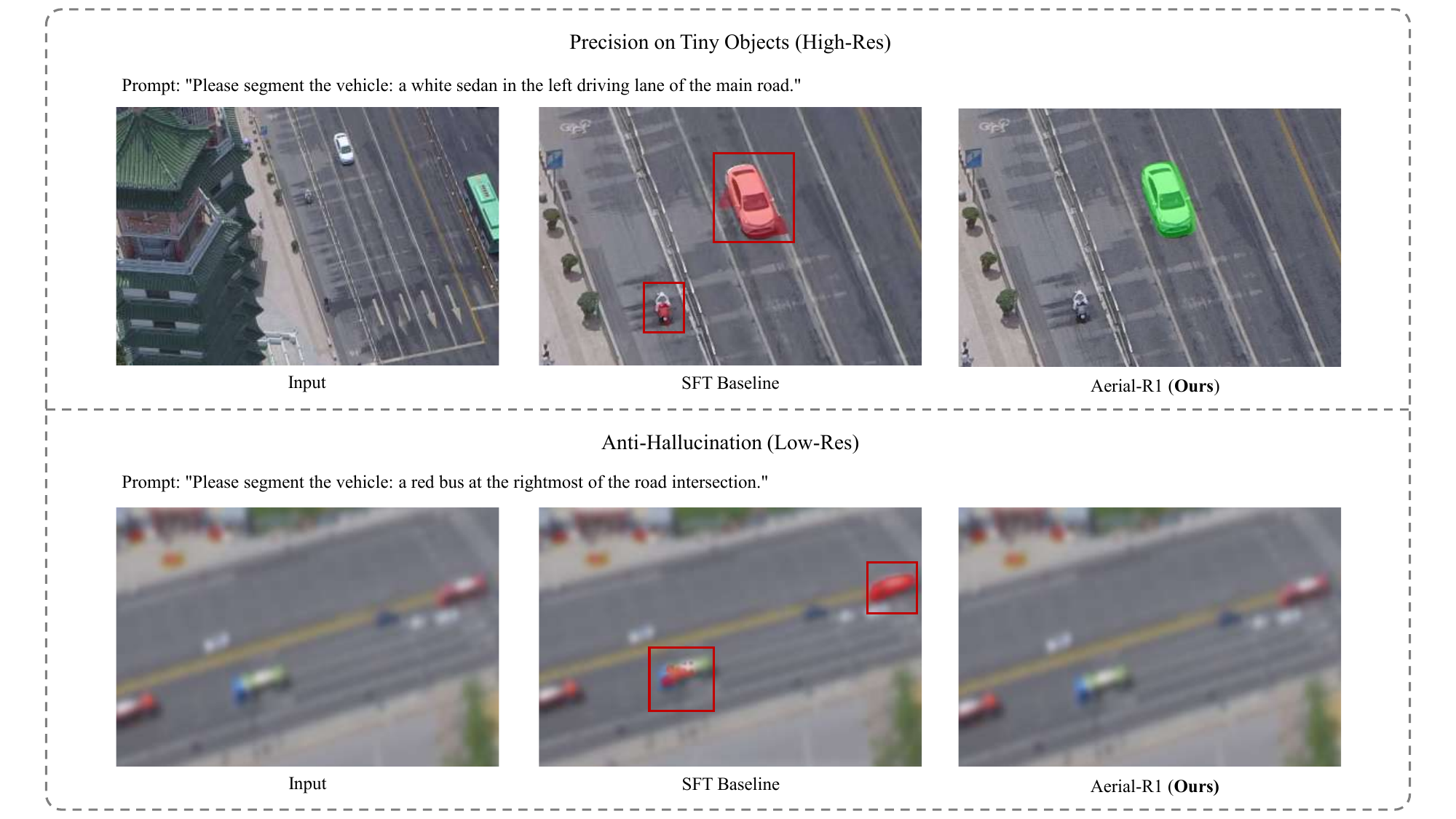}}
\caption{Qualitative comparison of the SFT baseline and Aerial-R1. The top row demonstrates precision on tiny objects in clear views, where the Sa2VA baseline exhibits over-segmentation and hallucinates masks for unrelated objects as highlighted by the red boxes. In contrast, Aerial-R1 generates a precise and singular mask for the target vehicle as highlighted by the green box. The bottom row showcases the anti-hallucination capability under visual degradation. While the baseline blindly hallucinates objects to satisfy the prompt as indicated by the red boxes, Aerial-R1 correctly identifies the lack of visual evidence and triggers the refusal mechanism by outputting an empty mask as indicated by the green box when visual evidence is insufficient.}
\label{fig:motivation}
\end{center}
\vskip -0.2in
\end{figure*}
Current approaches for aerial perception, including milestone foundation models like SkySense \cite{guo2024skysense} and recent large-scale multimodal models such as TerraMind \cite{terramind2025}, predominantly rely on Supervised Fine-Tuning (SFT). While effective for learning semantic mappings, SFT methods simply mimic the ground truth without modeling uncertainty. Consequently, when faced with tiny objects (often less than $32^2$ pixels) or visual ambiguity (e.g., motion blur, shadows), SFT models tend to exhibit two critical failures: (1) inconsistency characterized by unstable, coarse boundaries; and (2) hallucination where the model blindly predicts masks driven by language priors even when the target is unrecognizable. This phenomenon, known as object hallucination, is a persistent bottleneck in MLLMs \cite{li2023pope, zhang2023siren}. Although recent works have attempted to mitigate this via post-hoc correction \cite{yin2023woodpecker} or Reinforcement Learning from Human Feedback (RLHF) \cite{sun2023aligning, yu2024rlhfv}, these methods often require expensive annotated preference data or external detector modules, which are difficult to scale for dense aerial scenes.

We argue that these failures stem from a misalignment between \textit{memorization} and \textit{reasoning}. Specifically, current optimization objectives often encourage the model to hallucinate plausible answers to blindly satisfy user prompts, rather than grounding its responses in verifiable visual evidence. While SFT teaches the model \textit{what} the targets are, it fails to teach \textit{when} to predict.

Recent advances in LLM reasoning, notably Chain-of-Thought (CoT) \cite{wei2022chain} and Self-Consistency \cite{wang2023selfconsistency}, suggest that robust reasoning emerges from consistent logic paths. More recently, DeepSeek-R1 \cite{deepseek2024deepseekr1} demonstrated that Group Relative Policy Optimization (GRPO), originally introduced in DeepSeekMath \cite{shao2024deepseekmath}, can incentivize models to self-correct by enforcing consensus without a critic network. We posit a compelling parallel in the visual domain: Visual Consistency is the Chain-of-Thought of Vision. Just as a logical argument is robust only if it holds true under different phrasings, a visual perception is trustworthy only if it remains stable across varying resolutions. Inconsistency implies uncertainty, and uncertainty without refusal leads to hallucination. A trustworthy model should yield consistent pixel-level predictions across varying views (High-Res) and, crucially, maintain logical silence when visual evidence is corrupted (Low-Res).

To this end, we introduce Aerial-R1, the first framework to align aerial MLLMs for trustworthy perception via Hybrid-View Group Advantage-Weighted Regression (H-GAWR). A fundamental challenge in applying standard policy optimization (e.g., PPO or token-level GRPO) to segmentation tasks is that dense mask decoders operate non-autoregressively, making direct policy gradients mathematically intractable and highly unstable. To overcome this, H-GAWR adopts the Advantage-Weighted Regression (AWR) paradigm. We construct dynamic groups containing both High-Resolution (High-Res) and Low-Resolution (Low-Res) views to estimate a contrastive advantage without requiring a critic network. By using this group-relative advantage to weight the supervised segmentation objective, H-GAWR effectively performs stable policy extraction. By enforcing consensus on High-Res views and strictly penalizing hallucinations on Low-Res views, Aerial-R1 develops an explicitly incentivized refusal mechanism. Our main contributions are:
\begin{itemize}
    \item We propose \textbf{Aerial-R1}, the first aerial MLLM framework that shifts the paradigm from simple memorization to advantage-weighted visual reasoning, significantly improving the trustworthiness of aerial segmentation.
    \item We introduce \textbf{H-GAWR}, a novel optimization strategy that overcomes the limitations of token-level policy gradients by elegantly combining hybrid-view group sampling with Advantage-Weighted Regression. It optimizes for precision through a Consistency Reward and for safety via an Anti-Hallucination Penalty, without requiring manual negative annotations.
    \item We construct \textbf{VSAI-Ref}, the first comprehensive multimodal benchmark for the aerial domain that explicitly integrates a dedicated refusal testbed. Unlike existing positive-only datasets, it is specifically designed to rigorously evaluate both dense visual grounding and cognitive safety.
\end{itemize}

\section{Related Work}
\label{sec:related}

\subsection{Referring Image Segmentation in Aerial Imagery}
Referring Image Segmentation (RIS) aims to segment specific objects based on natural language queries. In the general domain, Large Multimodal Models (LMMs) have achieved remarkable success. Pioneering works like Shikra \cite{chen2023shikra} and Ferret \cite{you2024ferret} introduced referential dialogue capabilities, while LISA \cite{lai2024lisa} and Sa2VA \cite{yuan2025sa2va} further advanced dense reasoning segmentation. Furthermore, the integration of foundational vision models, particularly the Segment Anything Model (SAM) \cite{kirillov2023segment} and its successor SAM-2 \cite{ravi2024sam2}, has fundamentally shifted the paradigm of dense visual grounding. By bridging the reasoning capabilities of LLMs with the zero-shot promptable segmentation of SAM, recent architectures can decode complex linguistic queries into pixel-perfect masks. However, transferring these general-purpose models to the aerial domain presents significant challenges due to extreme scale variations and tiny objects, as benchmarked in foundational datasets like DOTA \cite{ding2018dota}, iSAID \cite{waqas2019isaid}, and VisDrone \cite{zhu2019visdrone}. Recent aerial-specific foundation models, such as SkySense \cite{guo2024skysense}, have attempted to bridge this gap via large-scale pre-training. Nevertheless, most existing approaches rely predominantly on Supervised Fine-Tuning (SFT). While SFT effectively aligns the semantic space, it simply mimics ground truth without modeling visual uncertainty. Consequently, when facing ambiguous tiny objects or complex backgrounds, SFT-based adaptation tends to produce coarse boundaries or severe hallucinations. Crucially, this hybrid architecture, where the MLLM operates autoregressively to generate functional tokens (e.g., \texttt{[SEG]}) while the mask decoder operates non-autoregressively, creates a distinct optimization gap. Traditional supervised objectives fail to dynamically align the semantic reasoning of the LLM with the continuous geometric output of the mask decoder. Aerial-R1 addresses this optimization gap by introducing an Advantage-Weighted Regression (AWR) paradigm to explicitly reward boundary precision and consistency without destabilizing the non-autoregressive mask generation.

\subsection{Hallucination Mitigation in MLLMs}
Object hallucination is a notorious bottleneck in MLLMs \cite{li2023pope, zhang2023siren}. This issue is exacerbated in aerial scenes where visual features are inherently weak or occluded. Prevailing mitigation strategies generally fall into two categories: (1) Post-hoc Correction, such as Woodpecker \cite{yin2023woodpecker}, which employs external detectors to verify outputs; and (2) RLHF Alignment, such as LLaVA-RLHF \cite{sun2023aligning} and RLHF-V \cite{yu2024rlhfv}, which optimize models using human or AI-annotated preference pairs. However, directly transplanting these prevailing strategies into the dense aerial segmentation domain encounters fundamental structural incompatibilities. 
First, post-hoc correction methods like Woodpecker heavily rely on frozen, off-the-shelf open-world object detectors (e.g., Grounding DINO) to verify visual evidence. In the aerial domain, where targets are typically extremely tiny ($<32^2$ pixels), dense, and highly ambiguous, these external general-domain detectors inherently fail. Relying on them introduces severe cascading errors rather than effective hallucination correction, making direct performance comparisons structurally inequitable.
Consequently, to ensure a rigorous and fair evaluation, we benchmark Aerial-R1 directly against Sa2VA \cite{yuan2025sa2va}, the state-of-the-art representative of the MLLM-SAM hybrid SFT paradigm.

\subsection{Reinforcement Learning for Reasoning}
Reinforcement Learning (RL) has become the standard for aligning LLMs with human intent. While PPO \cite{schulman2017proximal} is widely used, it requires an expensive value network (Critic). More recently, reference-free methods like DPO \cite{rafailov2023dpo} have gained popularity but still rely on paired data. To bridge this critical gap, we draw upon Advantage-Weighted Regression (AWR) \cite{peng2019advantage} and related offline RL paradigms like CRR \cite{wang2020critic}.
In the realm of logical reasoning, Chain-of-Thought (CoT) \cite{wei2022chain} and Self-Consistency \cite{wang2023selfconsistency} have demonstrated that correct reasoning paths tend to exhibit high consensus. Building on this, DeepSeek-R1 \cite{deepseek2024deepseekr1} recently proved that Group Relative Policy Optimization (GRPO) can significantly incentivize self-correction capabilities in LLMs via group-wise outcome comparison. However, standard policy gradient methods like GRPO are inherently designed for token-level autoregressive generation. Applying them directly to dense, non-autoregressive mask decoders is mathematically problematic; standard policy gradients applied to high-dimensional continuous pixel spaces suffer from extreme variance and unstable convergence. AWR elegantly reformulates the reinforcement learning problem into an advantage-weighted supervised learning objective. By restricting the policy update to remain close to the behavior policy via a Kullback-Leibler (KL) divergence constraint, AWR enables stable policy extraction directly from sampled trajectories without destabilizing the non-autoregressive decoder. Aerial-R1 pioneers the integration of AWR into the hybrid MLLM-SAM pipeline. By marrying the group-relative consensus sampling of GRPO with the optimization stability of AWR, our framework successfully distills abstract visual reasoning into precise geometric localization.

\section{Methodology}
\label{sec:method}

We present Aerial-R1, a framework designed to align MLLMs for aerial referring segmentation. The core innovation is \textbf{Hybrid-View Group Advantage-Weighted Regression (H-GAWR)}, a self-supervised reinforcement learning mechanism to enforce boundary consistency and suppress hallucinations in visually ambiguous scenarios.

\begin{figure*}[ht]
\vskip 0.15in
\begin{center}
\centerline{\includegraphics[trim=2cm 0cm 0cm 0cm, clip, width=1\textwidth]{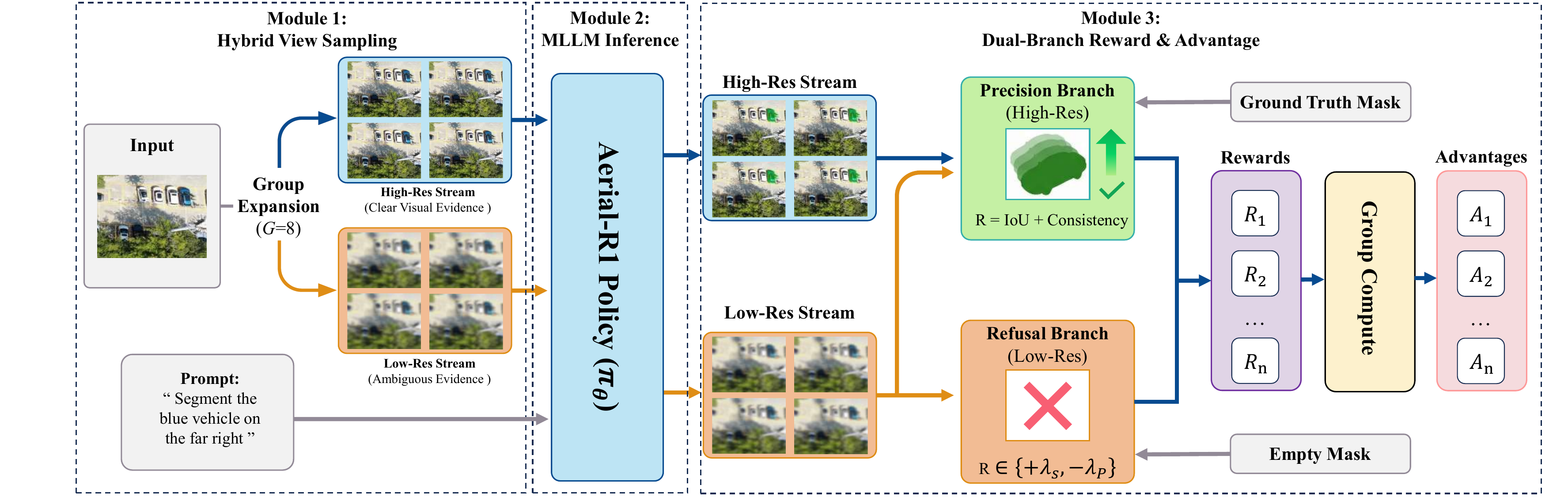}}
\caption{The architecture of Aerial-R1 with H-GAWR. Module 1 employs Hybrid View Sampling to construct a dynamic group ($G=8$) comprising both clear and degraded views. In Module 3, the Dual-Branch Reward mechanism assigns distinct objectives: the Precision Branch (top) rewards segmentation accuracy and Hybrid Consistency, enforcing alignment with the group consensus across both streams. Meanwhile, the Refusal Branch (bottom) cultivates safety by penalizing hallucinations ($-\lambda_p$) or rewarding silence ($+\lambda_s$) on ambiguous inputs. The final advantages are used to perform Advantage-Weighted Policy Extraction.}
\label{fig:framework}
\end{center}
\vskip -0.2in
\end{figure*}

\subsection{Preliminaries and Architecture}
\label{sec:preliminaries}

The proposed Aerial-R1 framework is designed to transition the MLLM from a passive pattern matcher to an active visual reasoner. To provide a robust foundation for this transition, {Aerial-R1 adopts the Sa2VA \cite{yuan2025sa2va} as its core architectural backbone}. The choice of Sa2VA is motivated by its unified and extensible design, which seamlessly integrates the high-level semantic reasoning of Large Language Models (LLMs) with the high-fidelity geometric localization of promptable segmentation models.

\paragraph{Multimodal Backbone and Adaptation}
The primary cognitive engine of our framework is based on the InternVL-Chat-V1.5 series \cite{chen2023internvl}, which possesses multi-scale visual processing capabilities essential for capturing the tiny, densely packed objects typical of aerial imagery. To bridge the domain gap between natural scenes and the ``God's eye view,'' we implement Parameter-Efficient Fine-Tuning (PEFT) via Low-Rank Adaptation (LoRA) \cite{hu2021lora}. Unlike standard fine-tuning, our LoRA matrices are strategically integrated into the \texttt{embed\_tokens} and \texttt{lm\_head} modules. This specific configuration allows the model to re-align its internal linguistic representations with aerial-specific semantics while preserving its general-purpose reasoning capacity. Formally, for a given input aerial image $I \in \mathbb{R}^{H \times W \times 3}$ and a language prompt $T$, the MLLM generates a hidden representation stream $\mathbf{H} = \text{MLLM}(I, T; \theta_{base}, \Delta\theta_{lora})$.

\paragraph{Dense Grounding Head: SAM-2 Integration}
For the mask generation module, we leverage the Segment Anything Model 2 (SAM-2) \cite{ravi2024sam2}. SAM-2's hierarchical feature memory and transformer-based mask decoder are pivotal for handling the extreme scale variations and complex background clutter in aerial datasets like VSAI-Ref. In the Aerial-R1 architecture, the SAM-2 vision encoder serves as the grounding encoder, while its non-autoregressive mask decoder generates the final binary mask $M \in \{0, 1\}^{H \times W}$. While the heavy vision encoder remains frozen to retain its zero-shot generalization, the mask decoder is kept unfrozen during the H-GAWR phase. This allows the decoder to adapt its continuous geometric output to the advantage-weighted signals derived from our reinforcement learning objective.

\paragraph{The Sa2VA Functional Interface}
The interaction between the MLLM and the grounding head is governed by the {Sa2VA functional interface}, which utilizes a special segmentation token, denoted as \texttt{[SEG]}. During the autoregressive generation process, the emergence of the \texttt{[SEG]} token acts as a semantic trigger. Let $\mathbf{h}_{seg} \in \mathbb{R}^{d_{llm}}$ be the hidden state of this token from the final LLM layer.
\begin{equation}
    \mathbf{z}_{grounding} = \text{Dropout}(\mathbf{W}_2 \cdot \text{ReLU}(\mathbf{W}_1 \cdot \mathbf{h}_{seg} + \mathbf{b}_1) + \mathbf{b}_2),
\end{equation}
where $\mathbf{W}_1$ and $\mathbf{W}_2$ are learnable weights. The resulting embedding $\mathbf{z}_{grounding}$ serves as a dynamic prompt that guides the SAM-2 decoder to perform conditional pixel-level localization: $M = \sigma(\text{Decoder}(\mathbf{z}_{grounding}, \mathcal{F}_{img}))$. By grounding Aerial-R1 in the Sa2VA architecture, we ensure that our H-GAWR strategy can focus on refining the model's reasoning logic and safety boundaries without needing to redesign the fundamental multimodal alignment.

\subsection{Hybrid-View Group Construction}
\label{sec:hybrid_view}

Standard Group Relative Policy Optimization (GRPO) samples multiple autoregressive trajectories from the exact same input to estimate policy gradients. However, applying this homogeneous sampling strategy to dense aerial imagery, which is inherently plagued by high visual ambiguity, often leads to premature mode collapse and redundant exploration in the continuous mask space. To overcome this, we propose \textbf{Hybrid-View Group Sampling}, a strategy designed to explicitly decouple semantic priors from structural evidence. 

For each training sample consisting of an image $I$ and a referring text $T$, we construct a diverse hybrid group $\mathcal{G}$ of size $G$:
\begin{equation}
    \mathcal{G} = \underbrace{\{I^{(1)}_{high}, \dots, I^{(G/2)}_{high}\}}_{\text{High-Res Views}} \cup \underbrace{\{I^{(1)}_{low}, \dots, I^{(G/2)}_{low}\}}_{\text{Degraded Views}},
\end{equation}
where the High-Res subset consists of identical copies of the pristine image $I$, retaining the original ground truth $M_{gt}$. The Low-Res subset consists of visually degraded variants generated via a continuous degradation operator $\mathcal{T}(\cdot)$, mapping the image to an information-depleted state while setting the corresponding ground truth to an empty set ($M_{gt} = \emptyset$).

\paragraph{Physical Motivation of Gaussian Degradation}
Rather than employing harsh downsampling (pixelation) or random noise injection, we formulate $\mathcal{T}$ as an isotropic Gaussian blur operator. Mathematically, it is defined as a spatial convolution $\mathcal{T}(I) = I * \mathcal{K}_{\sigma}$, where $\mathcal{K}_{\sigma}$ is a 2D Gaussian kernel (specifically configured with a kernel size of $15 \times 15$ and standard deviation $\sigma = 5.0$). This choice is driven by profound physical and representational rationales. In the context of extreme high-altitude aerial perception, target ambiguity does not naturally manifest as digital block artifacts. Instead, it predominantly arises from atmospheric scattering (e.g., Rayleigh and Mie scattering), sensor defocusing, and optical diffraction limits. These physical phenomena inherently act as natural low-pass filters in the spatial frequency domain.

\paragraph{Frequency-Domain Analysis and Cognitive Dissonance}
Representationally, aggressive downsampling (pixelation) violates the Nyquist-Shannon sampling theorem, introducing discrete, artificial spatial aliasing that unpredictably perturbs the continuous manifold of visual features. Conversely, Gaussian blur strictly attenuates high-frequency structural edges (object boundaries, fine textures) while perfectly preserving low-frequency global semantics (average chroma distributions, coarse illumination layouts). This smooth frequency attenuation accurately simulates realistic epistemic uncertainty in aerial environments, intentionally engineering a targeted \textit{cognitive dissonance} within the multimodal backbone.
The preserved low-frequency features suggest the target's presence, while the obliteration of high-frequency evidence leaves insufficient spatial support for the SAM-2 decoder.

By feeding this dissonance into our Advantage-Weighted objective, we force the policy to confront its perceptual limits. By penalizing hallucinated masks in information-depleted scenarios, we cultivate a decision boundary that balances recall with safety, echoing adversarial robustness \cite{madry2017towards} without external generative adversarial networks.

\subsection{Hybrid-View Group Advantage-Weighted Regression (H-GAWR)}
\label{sec:h_gawr}

To align the model's visual reasoning with the rigorous safety demands of aerial interpretation, we introduce \textbf{Hybrid-View Group Advantage-Weighted Regression (H-GAWR)}. While standard Reinforcement Learning from Human Feedback (RLHF) heavily relies on expensive, parameterized reward models (Critic), and token-level algorithms like Proximal Policy Optimization (PPO) or GRPO are primarily restricted to autoregressive token generation, applying these methods directly to high-dimensional, continuous mask decoders is mathematically intractable and notoriously unstable.

In contrast, H-GAWR bypasses the need for an external Critic network by leveraging intrinsic self-consistency. It elegantly bridges the reference-free, group-relative advantage estimation of GRPO with the mathematical stability of Advantage-Weighted Regression (AWR) \cite{peng2019advantage}. This creates a self-supervised RL paradigm explicitly tailored for dense, non-autoregressive mask prediction, framing policy optimization as an advantage-weighted structural loss minimization.

\subsubsection{Dual-Branch Reward Mechanism}
The core challenge in aerial perception is that optimizing solely for accuracy on clear targets inadvertently encourages aggressive guessing on ambiguous ones. To mitigate this, our reward function $r_i$ dynamically bifurcates based on the view type (High-Res or Low-Res). This dual-branch architecture decouples the learning objectives: the high-resolution stream focuses on exploiting structural boundaries (Precision), while the low-resolution stream focuses on calibrating cognitive uncertainty (Refusal).

\paragraph{Precision Branch (High-Res): Exploiting Consensus.} 
For high-resolution views $I \in \mathcal{G}_{high}$ containing valid targets, the objective is to achieve high geometric accuracy while maintaining robust reasoning consistency. Drawing inspiration from consistency regularization in semi-supervised learning \cite{tarvainen2017mean}, we define the reward for the $i$-th candidate mask $M_i$ as a composite metric:
\begin{equation}
    r_i = \text{IoU}(M_i, M_{gt}) + \beta \cdot \text{IoU}(M_i, \hat{M}_{cons}),
    \label{eq:reward_pos}
\end{equation}
where $\text{IoU}$ denotes the Intersection-over-Union \cite{everingham2010pascal}, and $\beta$ is a hyperparameter scaling the consensus reward. 

To explicitly enforce the \textit{Visual Chain-of-Thought}, we compute the consistency target by calculating the pixel-wise mean map of the \textit{entire} group's continuous logits: $\bar{P} = \frac{1}{G} \sum_{j=1}^G \sigma(\text{Logits}_j)$. We then derive the Robust Hybrid Consensus mask $\hat{M}_{cons}$ via binary thresholding:
\begin{equation}
    \hat{M}_{cons} = \mathbb{I}(\bar{P} > 0.5).
    \label{eq:cons_mask}
\end{equation}
\paragraph{Refusal Branch (Low-Res): Penalizing Hallucination.} 
Conversely, for the information-depleted views $I \in \mathcal{G}_{low}$ where the structural target is fully unrecognizable ($M_{gt} = \emptyset$), the fundamental objective shifts from localization to hallucination suppression. To enforce a strict cognitive boundary, we formulate a discrete reward structure based on the total predicted spatial area:
\begin{equation}
    r_i = \begin{cases} 
    +\lambda_{silence}, & \text{if } \sum M_i < \tau \quad (\text{successful refusal}), \\
    -\lambda_{penalty}, & \text{otherwise} \quad (\text{hallucination}),
    \end{cases}
    \label{eq:reward_neg}
\end{equation}
where $\sum M_i$ represents the aggregate number of foreground pixels predicted by the mask decoder. In our implementation, we define the noise tolerance threshold as $\tau=50$ pixels to account for negligible isolated activations. 

\subsubsection{Contrastive Advantage Estimation}
\begin{equation}
    A_i = \frac{r_i - \mu_G}{\sigma_G + \epsilon},
\end{equation}
where $\mu_G$ and $\sigma_G$ are the mean and standard deviation of the group rewards $\{r_1, \dots, r_G\}$, and $\epsilon = 10^{-8}$ ensures numerical stability. 

\begin{algorithm}[t]
   \caption{H-GAWR Training Process}
   \label{alg:h_gawr}
\begin{algorithmic}[1]
   \STATE {\bfseries Input:} Dataset $\mathcal{D}$, Policy $\pi_\theta$, Group Size $G$
   \STATE {\bfseries Hyperparams:} $\beta=2.0$, $\lambda_s=1.0$, $\lambda_p=2.0$, $\tau=50$
   \REPEAT
   \FOR{each batch $(I, T, M_{gt})$ in $\mathcal{D}$}
       \STATE  {\# 1. Hybrid-View Group Construction}
       \STATE $I_{high} \leftarrow \{I\} \times (G/2)$
       \STATE $I_{low} \leftarrow \{\mathcal{T}(I)\} \times (G/2)$
       \STATE $\mathcal{G} \leftarrow I_{high} \cup I_{low}$
       
       \STATE  {\# 2. Policy Rollout with Exploration}
        \STATE $\mathbf{z}_{grounding} \leftarrow f_{proj}(\mathbf{h}_{seg})$
        \IF{training}
            \STATE $\mathbf{z}_{grounding} \leftarrow \mathbf{z}_{grounding} + \epsilon \cdot (\|\mathbf{z}_{grounding}\|_2 \cdot \gamma)$ \COMMENT{Continuous Exploration}
        \ENDIF
        \STATE Sample $\{M_i\}_{i=1}^G \sim \text{Decoder}(\mathbf{z}_{grounding}, \mathcal{F}_{img})$
       
       \STATE  {\# 3. Dual-Branch Reward Calculation}
       \STATE Consensus $\hat{M}_{cons} \leftarrow \mathbb{I}\left(\frac{1}{G}\sum_{j=1}^G \sigma(M_j) > 0.5\right)$
       
       \FOR{$i=1$ to $G$}
           \IF{$i \in \text{High-Res}$}
               \STATE $r_i \leftarrow \text{IoU}(M_i, M_{gt}) + \beta \cdot \text{IoU}(M_i, \hat{M}_{cons})$
           \ELSE
               \STATE $r_i \leftarrow \begin{cases} +\lambda_s, & \text{if } \sum M_i < \tau \\ -\lambda_p, & \text{otherwise} \end{cases}$
           \ENDIF
       \ENDFOR
       
       \STATE  {\# 4. Advantage-Weighted Extraction}
       \STATE Compute Advantages $A_i \leftarrow (r_i - \mu_G) / (\sigma_G + \epsilon)$
       \STATE Isolate positive trajectories: $\mathcal{V} \leftarrow \{i \mid A_i > 0\}$
       \IF{$|\mathcal{V}| > 0$}
           \STATE Update $\theta$ by minimizing Eq. \ref{eq:total_loss}
       \ELSE
           \STATE Update $\theta$ using unweighted mean structural loss
       \ENDIF
   \ENDFOR
   \UNTIL{convergence}
\end{algorithmic}
\end{algorithm}

\subsubsection{Advantage-Weighted Policy Extraction}
Standard RL algorithms optimize the policy gradient via autoregressive log-probabilities. However, in our architecture, the semantic mask is generated by a non-autoregressive decoder (SAM-2). Applying standard policy gradients directly to dense, high-dimensional pixel outputs suffers from extreme variance. 

To propagate the reinforcement signal to the reasoning backbone without destabilizing the mask decoder, we formulate the optimization within the \textbf{Advantage-Weighted Regression (AWR)} \cite{peng2019advantage} framework. AWR reformulates RL as an advantage-weighted supervised learning problem. Instead of computing explicit policy gradients, we modulate the structural pixel-wise supervision using the estimated advantage. 

Specifically, we dynamically filter the group to retain only the "better-than-average" trajectories, denoted as the positive subset $\mathcal{V} = \{i \mid A_i > 0\}$. The total loss $\mathcal{L}_{total}$ combines the standard LLM log-likelihood with the advantage-weighted mask objective:
\begin{equation}
    \mathcal{L}_{total} = \mathcal{L}_{LLM} + \frac{1}{|\mathcal{V}| + \epsilon} \sum_{i \in \mathcal{V}} |A_i| \cdot \mathcal{L}_{mask}^{(i)},
    \label{eq:total_loss}
\end{equation}
where $\mathcal{L}_{mask}^{(i)}$ is specifically formulated to balance pixel-level classification and global region overlap, strictly defined as:
\begin{equation}
    \mathcal{L}_{mask}^{(i)} = 2.0 \cdot \mathcal{L}_{CE}(M_i, M_{gt}) + 0.5 \cdot \mathcal{L}_{Dice}(M_i, M_{gt}).
\end{equation}
In the rare event of severe mode collapse where no positive advantages exist ($|\mathcal{V}| = 0$), the objective safely falls back to the unweighted mean of $\mathcal{L}_{mask}^{(i)}$ to maintain gradient flow.

\subsubsection{Theoretical Justification of H-GAWR}
Let $x = (I, T)$ and $y = M$.
\begin{equation}
\begin{aligned}
    \max_{\pi} \quad & \mathbb{E}_{x \sim \mathcal{D}, y \sim \pi(\cdot|x)} \left[ A(x,y) \right] \\
    \text{s.t.} \quad & \mathbb{E}_{x \sim \mathcal{D}} \left[ D_{\text{KL}}(\pi(\cdot|x) \parallel \pi_{ref}(\cdot|x)) \right] \le \delta,
\end{aligned}
\end{equation}
where $D_{\text{KL}}$ is the Kullback-Leibler divergence. Using a Lagrange multiplier $\alpha > 0$, the closed-form optimal non-parametric policy $\pi^*$ is derived as:
\begin{equation}
    \pi^*(y|x) = \frac{1}{Z(x)} \pi_{ref}(y|x) \exp\left( \frac{A(x,y)}{\alpha} \right).
    \label{eq:optimal_policy}
\end{equation}
We project this optimal policy onto our neural network $\pi_\theta$ by minimizing the KL divergence $D_{\text{KL}}(\pi^* \parallel \pi_\theta)$, which is mathematically equivalent to solving the weighted Maximum Likelihood Estimation (MLE):
\begin{equation}
    \arg\max_\theta \mathbb{E}_{x \sim \mathcal{D}, y \sim \pi_{ref}} \left[ \exp\left( \frac{A(x,y)}{\alpha} \right) \log \pi_\theta(y|x) \right].
    \label{eq:awr_projection}
\end{equation}

\subsubsection{Continuous Embedding Exploration}
To further enhance the exploration efficiency in the continuous geometric space of the non-autoregressive SAM-2 decoder, we introduce a \textit{Continuous Temperature Sampling} mechanism during the training phase. Specifically, instead of directly passing the grounding embedding $\mathbf{z}_{grounding}$ to the mask decoder, we perturb it with scaled stochastic noise:
\begin{equation}
    \tilde{\mathbf{z}}_{grounding} = \mathbf{z}_{grounding} + \epsilon \cdot (\|\mathbf{z}_{grounding}\|_2 \cdot \gamma),
\end{equation}
where $\epsilon \sim \mathcal{N}(0, \mathbf{I})$ represents a standard Gaussian noise vector, and $\gamma$ is a temperature hyperparameter that controls the exploration intensity. This perturbation forces the policy to remain robust to minor semantic fluctuations and prevents the non-autoregressive head from collapsing into sub-optimal geometric modes.

\section{The VSAI-Ref Benchmark}
\label{sec:benchmark}

While established aerial datasets such as DOTA \cite{ding2018dota} and iSAID \cite{waqas2019isaid} have profoundly advanced closed-set object detection, they exhibit a fundamental structural limitation when evaluating Multimodal Large Language Models (MLLMs): they are intrinsically \textit{positive-only} and \textit{object-centric}. In these traditional benchmarks, every evaluation metric assumes the target is physically present. Consequently, they inherently fail to assess a model's susceptibility to \textit{language-driven hallucinations}, the severe failure mode where an MLLM blindly fabricates a prediction merely to satisfy a user's prompt, even when the visual evidence is entirely absent or severely corrupted. Evaluating the true trustworthiness of an aerial agent requires probing its cognitive safety boundary: its capacity to explicitly refuse an unreasonable query. Because traditional datasets lack diverse linguistic referring expressions paired with strictly verified \textit{hard negative} scenarios, utilizing them to benchmark MLLM hallucination and safety alignment is structurally invalid. To bridge this critical methodological gap, we construct VSAI-Ref. Importantly, VSAI-Ref is not a synthesized toy dataset; it strictly inherits the authentic high-altitude complexity, sensor characteristics, and scale variations of the original VSAI dataset \cite{wang2022vsai}. By introducing a rigorous multi-task linguistic framework equipped with dedicated refusal testbeds, VSAI-Ref stands as the first scientifically robust benchmark to jointly assess dense grounding precision and hallucination robustness in the aerial domain, fundamentally shifting the evaluation paradigm from simple object recall to trustworthy cognitive safety. 
Furthermore, to address concerns regarding cross-dataset generalization, VSAI-Ref intrinsically preserves the massive sensor heterogeneity and extreme environmental diversity (e.g., varying flight altitudes, distinct UAV platforms, fog, and nighttime conditions) of the foundational VSAI dataset \cite{wang2022vsai}. This inherent structural diversity ensures that the benchmark rigorously evaluates true zero-shot generalization across different optical sensors and adverse atmospheric conditions, rather than mere dataset-specific overfitting.

\subsection{Pipeline Construction via MLLMs} 
Constructing large-scale referring expressions for dense aerial scenes is labor-intensive. We propose an automated and scalable pipeline leveraging Qwen-VL-Max to generate high-quality annotations from the original VSAI dataset \cite{wang2022vsai}. For referring image segmentation, we extract visual crops for each valid vehicle instance and prompt the model to generate unique referring expressions, such as \textit{a white sedan parked next to a tree}. To ensure high-quality and distinct annotations, we instruct the model to act as an aerial image analyst with strict constraints: (1) Describe ONLY the target vehicle inside the marked area; (2) Do NOT mention the environment (e.g., roads, trees) unless necessary for disambiguation; (3) Include position and appearance attributes (e.g., color, type); and (4) Keep the description concise ($<$ 15 words). We then filter out ambiguous descriptions that could apply to multiple targets in the same image.

To evaluate generalized reasoning through visual question answering, we generate QA pairs focusing on attributes like color and type in addition to spatial relations. We enforce a structured JSON format (\texttt{\{"question": "string", "answer": "string"\}}) by instructing the model to generate objective questions based solely on visual content and to strictly avoid simple 'Yes/No' queries. Additionally, to assess holistic contextual understanding, we generate 214 comprehensive \textit{scene description} samples using the same Qwen-VL-Max pipeline. These samples require the model to narrate the global context, spatial distributions, and inter-object relationships within the entire aerial view. Furthermore, we unify the input format across different tasks during training and evaluation using a standard chat template. For specific vehicle segmentation, the prompt is structured as \texttt{"User: Please segment the specific vehicle: \{description\} Assistant: Sure, [SEG]."}, where the \texttt{[SEG]} token triggers the mask decoder. General vehicle segmentation uses the prompt \texttt{"User: Please segment all vehicles. Assistant: Sure, [SEG]."}. For free-form reasoning, the VQA template is defined as \texttt{"User: \{question\} Assistant: \{answer\}"}, and the global scene understanding is prompted via \texttt{"User: Please describe this image in detail. Assistant: \{scene\_description\}"}.

\subsection{The Hard Negative Subset}
\label{sec:hard_negatives}
To rigorously evaluate model trustworthiness, VSAI-Ref explicitly includes a \textit{Hard Negative} subset. 
Unlike standard benchmarks where every referring expression corresponds to a present object, we integrate the \textit{non-vehicle image subset} from the original VSAI dataset\cite{wang2022vsai}, resulting in a challenging test set of 832 negative samples. These images depict pure backgrounds (e.g., dense forests, rivers, or empty roads) devoid of any vehicle instances. 
To construct the linguistic queries for these negative images, we employ a context-aware adversarial prompt strategy. We manually curate a specialized pool of 13 diverse, error-prone vehicle descriptions (encompassing various colors, types, and spatial relations). For each empty background image, we instruct Qwen-VL-Max to act as an adversarial evaluator: it analyzes the scene context and selects the most deceptive prompt from the 13-item pool, specifically choosing a target that logically \textit{could} exist in that specific environment but is completely absent. This setup establishes a rigorous, contextually deceptive testbed for explicit refusal, evaluating whether the model has successfully learned the safety boundary to resist hallucinating objects driven by plausible language priors.

\subsection{Dataset Statistics}

\begin{figure}[h]
\centering
\includegraphics[width=\linewidth]{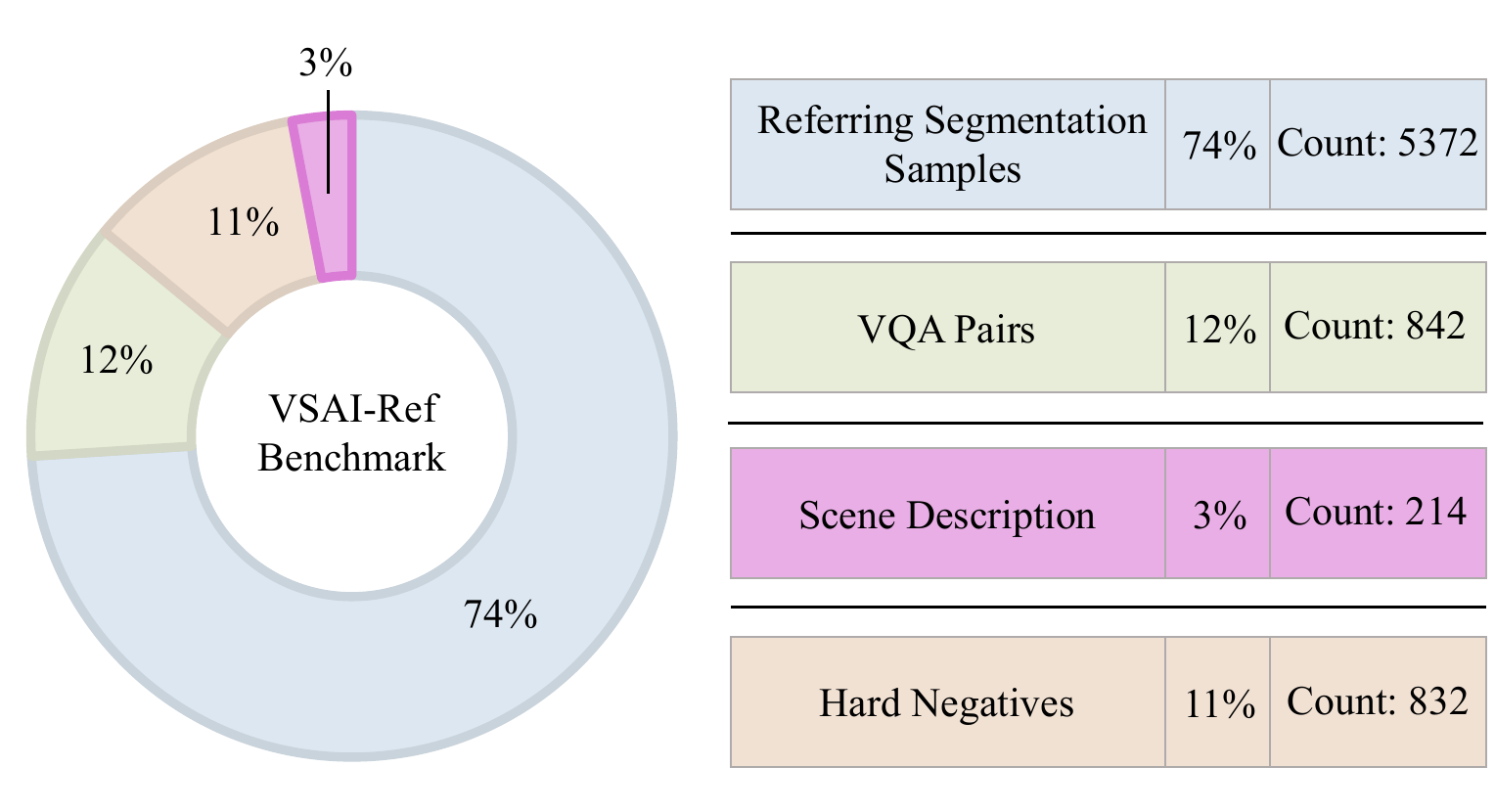} 
\caption{Data distribution of the VSAI-Ref benchmark. The proportions of referring segmentation samples, VQA pairs, scene descriptions, and hard negatives reflect the comprehensive multi-task nature of the dataset.}
\label{fig:data_dist}
\vskip -0in
\end{figure}

 The VSAI-Ref benchmark is structured to verify both grounding precision and hallucination robustness. It comprises 5,372 positive samples for standard referring segmentation, strictly partitioned into training, validation, and testing sets. Complementing these are the 832 hard negative samples, derived exclusively from VSAI \cite{wang2022vsai} images containing no vehicles, which are reserved specifically for testing. As illustrated in Figure \ref{fig:data_dist}, the dataset employs a multi-task annotation strategy where referring segmentation samples, VQA pairs, and negative samples are integrated to evaluate the model's comprehensive understanding. The detailed numerical distribution across different splits is summarized in Table \ref{tab:dataset_stats_main}.

\begin{table}[h]
\caption{Statistics of the VSAI-Ref benchmark. The dataset integrates multiple reasoning tasks, including a newly added scene description subset and a dedicated hard negative subset of 832 adversarially selected samples to rigorously evaluate hallucination.}
\label{tab:dataset_stats_main}
\begin{center}
\begin{small}
\begin{sc}
\resizebox{\linewidth}{!}{
\begin{tabular}{l|ccc|c} 
\toprule
Split & Ref. Seg. Samples & VQA Pairs & Scene Desc. & Hard Negatives \\ 
\midrule
Training & 3,003 & - & - & - \\
Validation & 886 & - & - & - \\
Testing & 1,483 & 842 & 214 & 832 \\ 
\midrule
Total & 5,372 & 842 & 214 & 832 \\
\bottomrule
\end{tabular}
}
\end{sc}
\end{small}
\end{center}
\vskip -0.1in
\end{table}

\section{Experiments}
\label{sec:experiments}

\subsection{Experimental Setup}
\label{sec:setup}

\textbf{Datasets.} 
We evaluate our method on the VSAI-Ref benchmark. As detailed in Section \ref{sec:benchmark}, this dataset is constructed to require dense, attribute-aware reasoning. It includes a specific Hard Negative subset to rigorously benchmark the model's robustness against hallucination.

\textbf{Implementation Details.}
We implement Aerial-R1 using PyTorch \cite{paszke2019pytorch} and DeepSpeed \cite{rasley2020deepspeed} on 4 NVIDIA H100 GPUs with gradient accumulation to simulate a larger effective batch size. The policy $\pi_\theta$ is initialized from the Sa2VA architecture with an InternVL-Chat-V1-5 visual backbone (frozen) and a SAM-2 mask decoder (unfrozen, initialized from pre-trained weights) \cite{yuan2025sa2va}. Input images are resized to $1024 \times 1024$. To maintain the general reasoning capability of the pre-trained MLLM while adapting to the aerial domain, we employ Low-Rank Adaptation (LoRA) \cite{hu2021lora} on the \texttt{embed\_tokens} and \texttt{lm\_head} modules with Rank $r=128$, Alpha $\alpha=256$, and a dropout rate of 0.05. We use a relatively high rank to accommodate the complex spatial reasoning required for pixel-level grounding. During the SFT stage, we apply a dataset repetition factor of 10 within the single epoch to ensure sufficient semantic alignment. In contrast, the dataset is iterated only once during the H-GAWR stage to prevent over-optimization.

The training consists of two phases: Supervised Fine-Tuning (SFT) and Hybrid-View Group Advantage-Weighted Regression (H-GAWR). Optimization uses AdamW \cite{loshchilov2017decoupled} with a cosine scheduler. The detailed hyperparameters for both training stages are comprehensively summarized in Table \ref{tab:hyperparams}. Notably, we deliberately reduced the learning rate from $4e^{-5}$ in the SFT stage to $1e^{-6}$ during the H-GAWR stage to prevent policy collapse, as the advantage-weighted extraction is more sensitive to gradient updates. The Hybrid-View Group size is set to $G=8$, with consistency coefficient $\beta=2.0$, hallucination penalty $\lambda_p=2.0$, and silence reward $\lambda_s=1.0$. During the H-GAWR phase, we set the exploration temperature $\gamma = 0.1$ to facilitate feature-level diversity within each hybrid-view group. This stochastic perturbation is only applied during training to ensure deterministic inference.

\textbf{Computational Overhead and Deployment Efficiency.} 
A critical consideration in scaling reinforcement learning for dense visual tasks is the associated computational and engineering complexity. We acknowledge that the proposed H-GAWR introduces additional computational overhead during the training phase compared to standard SFT. Specifically, constructing the hybrid-view group of size $G=8$ and computing the contrastive advantages necessitates multiple forward passes, which increases the memory footprint and training time per optimization step. However, this cost is highly manageable and mitigated by our gradient accumulation strategy and the sample efficiency of AWR, which achieves convergence in merely a single epoch. 
Crucially, this structural complexity is strictly confined to the training phase. During inference and real-world deployment, the Aerial-R1 architecture reverts completely to the standard single-pass Sa2VA pipeline. Unlike post-hoc correction frameworks (e.g., Woodpecker \cite{yin2023woodpecker}) that rely heavily on slow, external open-world object detectors for verification, Aerial-R1 requires no multi-view sampling, no advantage estimation, and no external modules at test time. Consequently, our method introduces {strictly zero additional latency or memory overhead} during inference compared to the base SFT model. This fundamental decoupling of training complexity from inference efficiency makes Aerial-R1 highly practical for real-time, resource-constrained autonomous aerial agents.

\begin{table*}[t]
\caption{Detailed Hyperparameter Settings for Aerial-R1 Training Stages.}
\label{tab:hyperparams}
\begin{center}
\begin{small}
\begin{sc}
\begin{tabular*}{\textwidth}{@{\extracolsep{\fill}}llcc}
\toprule
Category & Hyperparameter & SFT Stage & H-GAWR Stage \\
\midrule
\multirow{8}{*}{\textit{Optimization Config}} 
& Optimizer & AdamW & AdamW \\
& Learning Rate & $4e^{-5}$ & $1e^{-6}$ \\
& Weight Decay & 0.05 & 0.05 \\
& LR Scheduler & Cosine w/ Warmup & Cosine w/ Warmup \\
& Warmup Ratio & 0.05 & 0.05 \\
& Batch Size (per GPU) & 4 & 2 \\
& Gradient Accumulation & 32 & 32 \\
& Max Epochs & 1 & 1 \\
\midrule
\multirow{6}{*}{\textit{H-GAWR Specifics}} 
& Group Size ($G$) & -- & 8 \\
& Consistency Beta ($\beta$) & -- & 2.0 \\
& Hallucination Penalty ($\lambda_{p}$) & -- & 2.0 \\
& Silence Reward ($\lambda_{s}$) & -- & 1.0 \\
& Exploration Temperature ($\gamma$) & -- & 0.1 \\
& Gaussian Blur (Default) & -- & Kernel $15 \times 15$, $\sigma=5.0$ \\
& Pixelation (Ablation) & -- & Downsample Ratio $0.125$ \\
& Gaussian Noise (Ablation) & -- & Standard Deviation $\sigma_{n}=0.3$ \\
& Noise Tolerance ($\tau$) & -- & 50 pixels \\
\bottomrule
\end{tabular*}
\end{sc}
\end{small}
\end{center}
\vskip -0.1in
\end{table*}

\subsection{Main Results}
\label{sec:main_results}

We evaluate Aerial-R1 using mIoU, Precision, Recall, and F1-Score, with results summarized in Table \ref{tab:main_results}. To rigorously validate our advantage-weighted learning paradigm, we incorporate a straightforward data augmentation baseline, denoted as Sa2VA (SFT + Low-Res Aug.). This baseline randomly replaces high-resolution inputs with low-resolution images (specifically utilizing pixelation downsampling) and empty ground truths during standard supervised fine-tuning, attempting to force the model to memorize the refusal behavior.

Regarding the analysis on Task 1, Aerial-R1 consistently outperforms both baselines. Compared to the standard SFT, Aerial-R1 (Noise) achieves a substantial improvement in mIoU (+3.16\%) and F1-Score (+3.13\%). Crucially, the naive Sa2VA (SFT + Low-Res Aug.) baseline actively degrades performance across all metrics (mIoU dropping from 42.96\% to 40.83\%). This phenomenon confirms that passively forcing the model to memorize empty masks under ambiguity disrupts its semantic alignment and exacerbates uncertainty. In contrast, the negative reward branch of H-GAWR effectively cultivates a conservative safety boundary without destabilizing the feature space. In Task 2, Aerial-R1 (Blur) achieves massive gains, notably boosting mIoU to 57.53\% and Recall to 68.86\% (+2.75\% over SFT). Enforcing consensus between high-resolution and degraded views aids the model in discovering harder, smaller, or partially occluded targets. Furthermore, Aerial-R1 exhibits highly robust performance regardless of the specific degradation type used during training (Blur, Pixelation, or Noise). All three variants achieve strictly comparable and superior results against the baselines, indicating that the advantage-weighted extraction successfully learns a generalized reasoning capability rather than overfitting to a specific artificial noise pattern.

\begin{table}[h]
\caption{Quantitative Comparison on VSAI-Ref. Aerial-R1 achieves the best overall trade-off (mIoU/F1) across both tasks. Compared to the standard SFT and naive augmentation baselines, our method equipped with different degradation variants (Blur, Pixelation, Noise) consistently achieves significant gains.}
\label{tab:main_results}
\begin{center}
\begin{small}
\begin{sc}
\resizebox{\linewidth}{!}{
\begin{tabular}{l|cccc}
\toprule
Method & mIoU & Prec. & Recall & F1 \\
\midrule
\multicolumn{5}{c}{\textit{Task 1: Specific Vehicle Segmentation}} \\
\midrule
Sa2VA (SFT) \cite{yuan2025sa2va} & 42.96 & 51.93 & 49.53 & 49.66 \\
Sa2VA (SFT + Low-Res Aug.) & 40.83 & 50.22 & 47.67 & 47.66 \\
\midrule
\textbf{Aerial-R1 (Blur)} & 46.06 & 54.91 & 52.36 & 52.76 \\
\textbf{Aerial-R1 (Pixelation)} & 46.07 & \textbf{55.01} & 52.18 & 52.71 \\
\textbf{Aerial-R1 (Noise)} & \textbf{46.12} & 54.87 & \textbf{52.44} & \textbf{52.79} \\
\midrule
\multicolumn{5}{c}{\textit{Task 2: General Vehicle Segmentation}} \\
\midrule
Sa2VA (SFT) \cite{yuan2025sa2va} & 54.93 & 73.20 & 66.11 & 67.62 \\
Sa2VA (SFT + Low-Res Aug.) & 53.56 & 71.70 & 65.33 & 66.30 \\
\midrule
\textbf{Aerial-R1 (Blur)} & \textbf{57.53} & 74.52 & \textbf{68.86} & \textbf{69.93} \\
\textbf{Aerial-R1 (Pixelation)} & 57.45 & \textbf{75.05} & 68.30 & 69.85 \\
\textbf{Aerial-R1 (Noise)} & 57.52 & 74.50 & \textbf{68.86} & 69.92 \\
\bottomrule
\end{tabular}
}
\end{sc}
\end{small}
\end{center}
\vskip -0.1in
\end{table}

\begin{figure*}[t]
\vskip 0.1in
\begin{center}
\centerline{\includegraphics[width=0.98\textwidth]{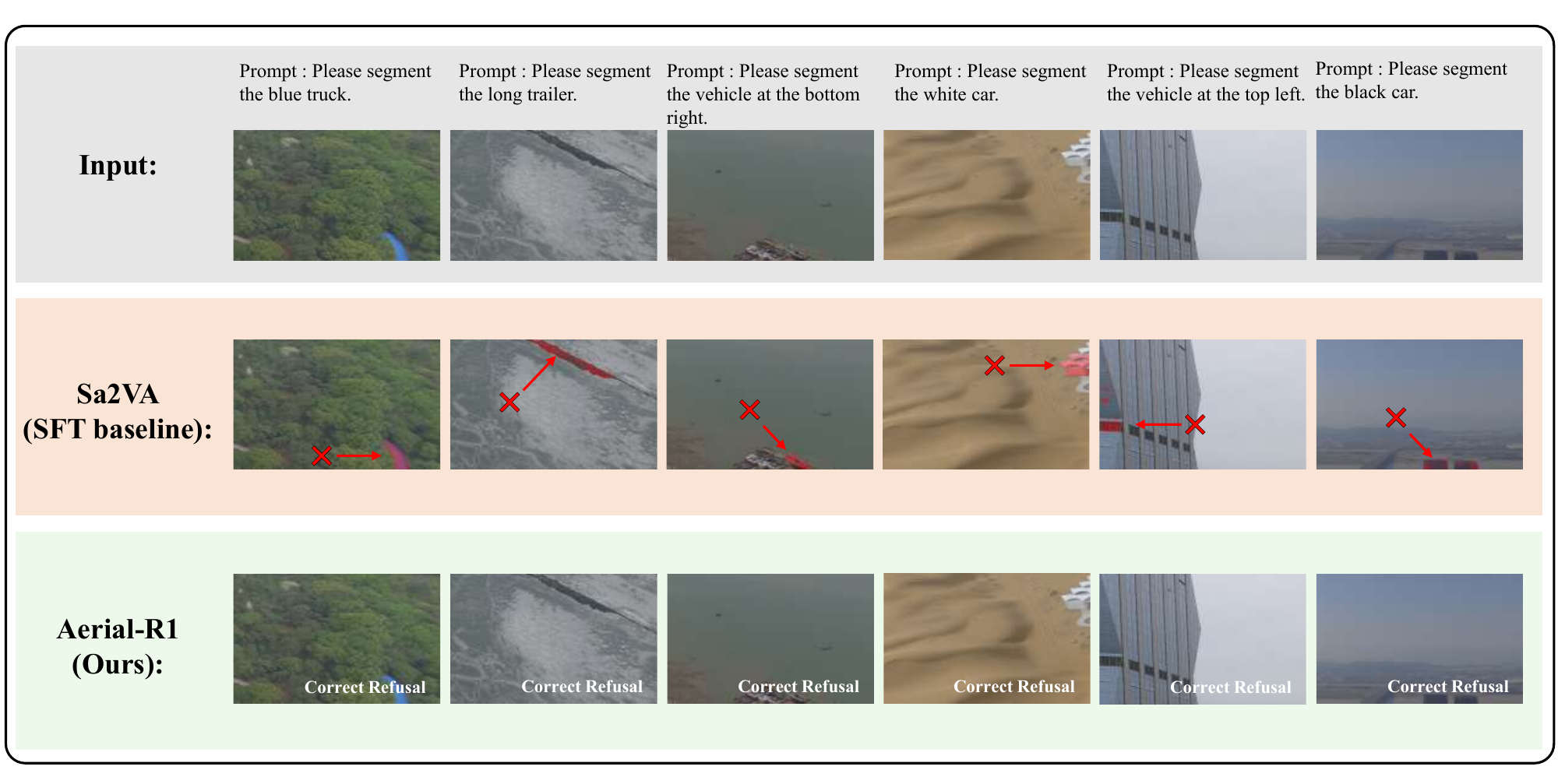}}
\caption{Qualitative comparison on Hard Negatives. We visualize challenging scenarios, including dense forest and hazy water, where the target objects described in the prompts are visually unrecognizable or absent. (Middle Row) The SFT baseline fails to handle uncertainty, generating severe hallucinations in red masks driven by language priors. (Bottom Row) Aerial-R1 correctly identifies the lack of visual evidence and triggers the explicitly incentivized refusal mechanism by outputting empty masks, demonstrating superior trustworthiness.}
\label{fig:robustness}
\end{center}
\vskip -0.2in
\end{figure*}

\subsection{Robustness Analysis}
\label{sec:robustness}

\begin{figure*}[t!] 
    \centering
    \includegraphics[width=1.4\linewidth, height=0.35\textheight, keepaspectratio]{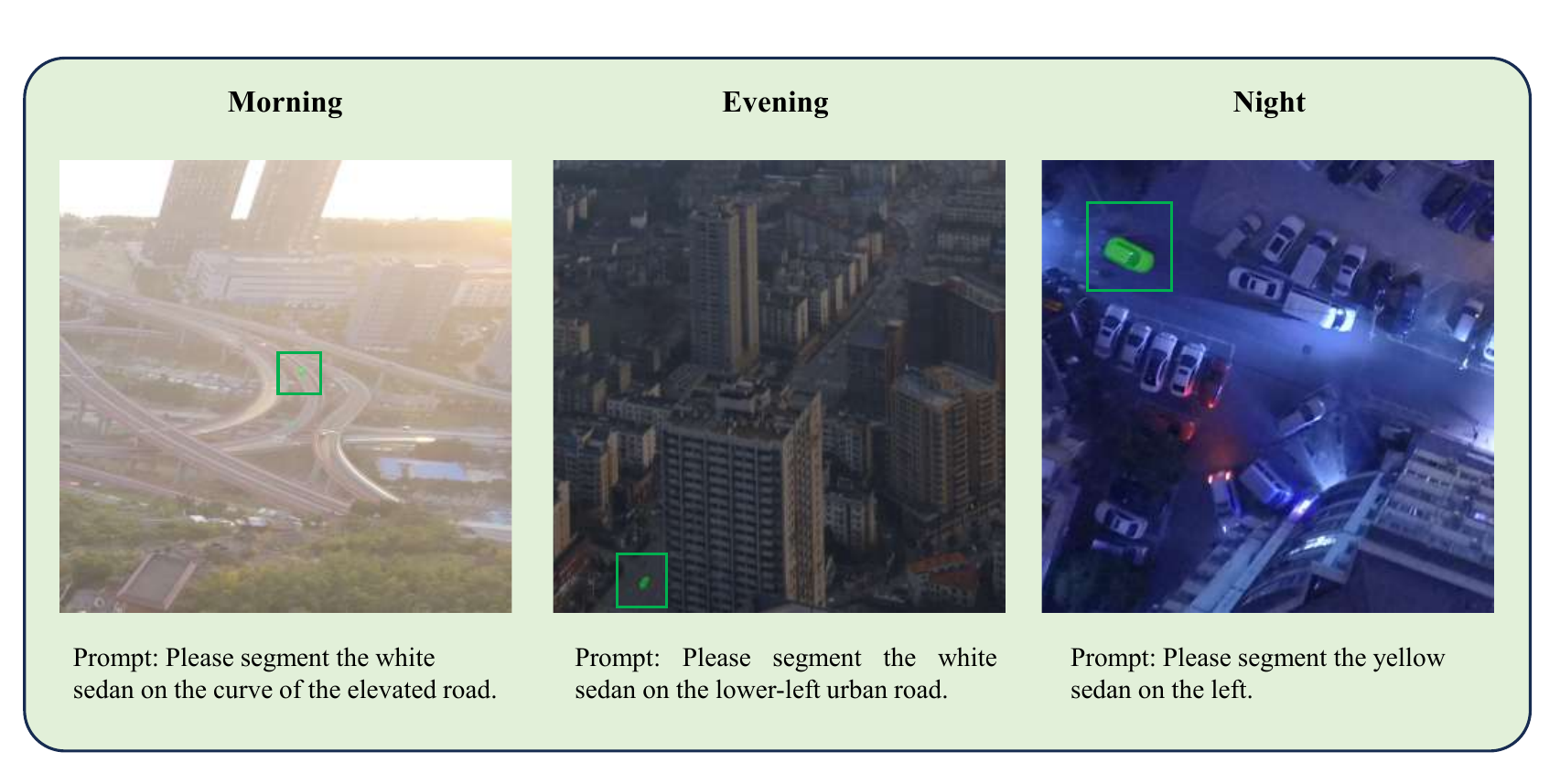}
    \centerline{\small (a) Performance of Aerial-R1 across different spatial scenarios.}
    \includegraphics[width=1.4\linewidth, height=0.35\textheight, keepaspectratio]{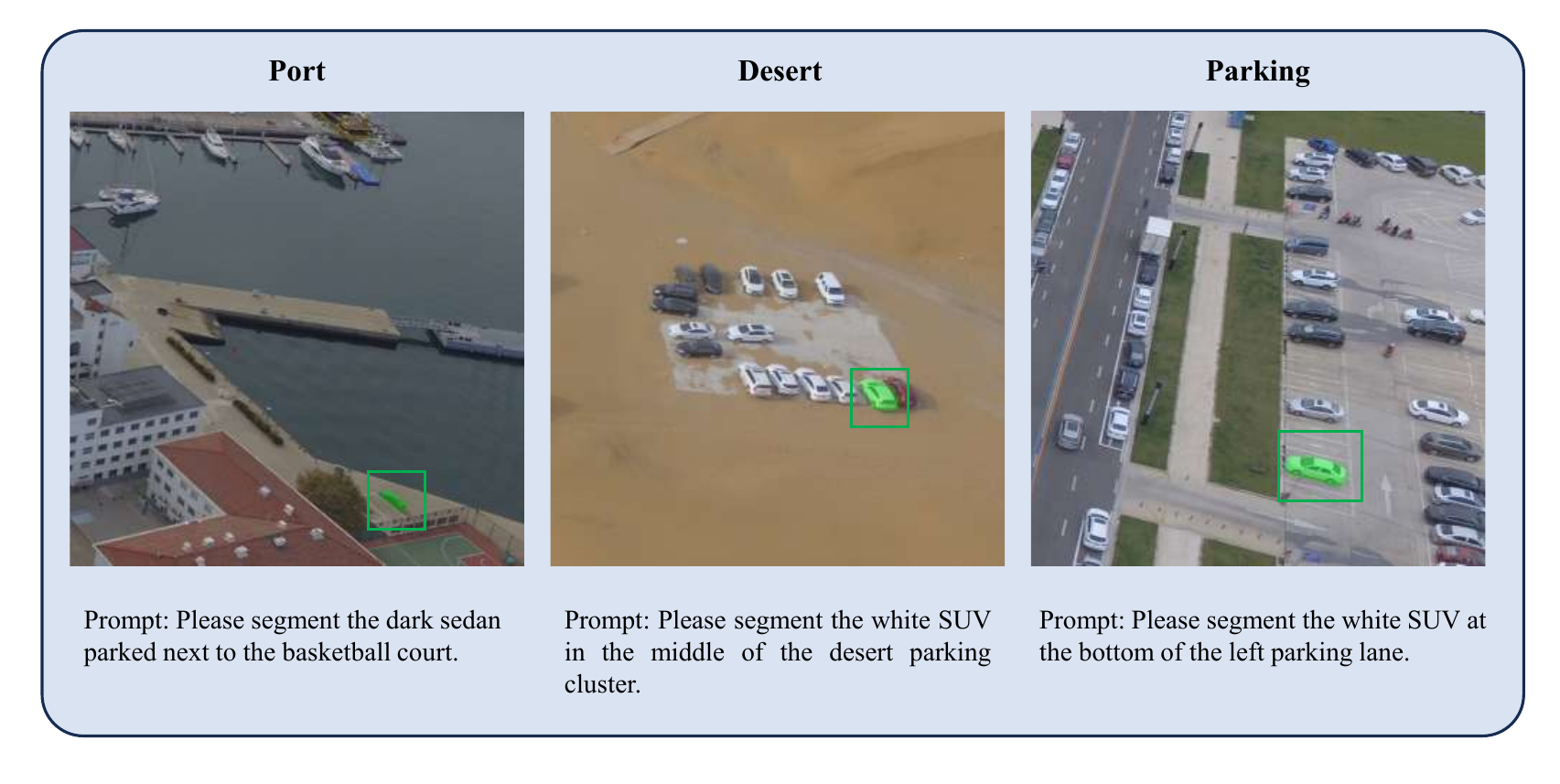}
    \centerline{\small (b) Performance of Aerial-R1 under varying temporal (lighting) conditions.}
    \caption{Robustness Visualization Results. (a) Performance of Aerial-R1 across different spatial scenarios. (b) Performance of Aerial-R1 under varying temporal (lighting) conditions. The model's predictions are highlighted in green boxes.}
    \label{fig:robustness_vis}
\end{figure*}

We define a hard negative subset consisting of 832 ambiguous or empty samples to rigorously evaluate the models' capacity for explicit refusal. Table \ref{tab:rescue_rate} quantifies the performance against language-driven hallucinations. The standard Sa2VA (SFT) struggles significantly, successfully refusing only 38.82\% of the deceptive queries while hallucinating on the remaining 509 samples. While the naive data augmentation baseline, Sa2VA (SFT + Low-Res Aug.), improves the pass rate to 52.64\% by memorizing empty masks under ambiguity, this comes at the severe cost of degrading normal segmentation performance as previously established in Table \ref{tab:main_results}. In contrast, our H-GAWR framework achieves the highest successful refusal rate of 61.18\%. By explicitly penalizing hallucinations via advantage-weighted regression without corrupting the fundamental semantic alignment, Aerial-R1 successfully silences a significant portion of the baseline's false positives, demonstrating a vastly superior and trustworthy safety boundary. To further demonstrate the robust generalization of Aerial-R1, we provide additional qualitative visualizations. As depicted in Figure \ref{fig:robustness_vis} (a), the model maintains high precision in diverse spatial scenarios, including ports, deserts, and dense parking lots. Furthermore, as illustrated in Figure \ref{fig:robustness_vis} (b), the model effectively adapts to varying temporal lighting conditions, achieving stable segmentation from morning haze to nighttime darkness.

\begin{table}[h]
\caption{Performance on Hard Negatives (832 samples). Aerial-R1 achieves the highest successful refusal rate, significantly mitigating language-driven hallucinations compared to the SFT baseline and naive augmentation strategies.}
\label{tab:rescue_rate}
\begin{center}
\begin{small}
\begin{sc}
\resizebox{\columnwidth}{!}{
\begin{tabular}{l|ccc}
\toprule
Method & Refused (\checkmark) & Hallucinated ($\times$) & Pass Rate \\
\midrule
Sa2VA (SFT) & 323 & 509 & 38.82\% \\
Sa2VA (SFT + Low-Res Aug.) & 438 & 394 & 52.64\% \\
\textbf{Aerial-R1 (Ours)} & \textbf{509} & \textbf{323} & \textbf{61.18\%} \\
\bottomrule
\end{tabular}
}
\end{sc}
\end{small}
\end{center}
\vskip -0.1in
\end{table}

\subsection{Ablation Study}
\label{sec:ablation}

To verify the contribution of each component in the H-GAWR framework, we conduct ablation studies on both the Specific Vehicle Segmentation and General Vehicle Segmentation tasks. We specifically analyze the impact of the Consistency Reward for boundary refinement and the Low-Res Stream for explicitly incentivized refusal.

\begin{table}[htb]
\caption{Ablation Study on VSAI-Ref. We investigate the impact of removing key components across both tasks. The Aerial-R1 (Full) denotes the model trained with the complete H-GAWR strategy (utilizing the default Gaussian Blur).}
\label{tab:ablation}
\begin{center}
\begin{small}
\begin{sc}
\resizebox{\linewidth}{!}{
\begin{tabular}{l|ccc|c}
\toprule
Variant & mIoU & Prec. & Recall & $\Delta$ mIoU \\
\midrule
\multicolumn{5}{c}{\textit{Task 1: Specific Vehicle Segmentation}} \\
\midrule
\textbf{Aerial-R1 (Full)} & \textbf{46.06} & \textbf{54.91} & \textbf{52.36} & - \\
(1) w/o Consistency & 44.41 & 53.33 & 50.82 & -1.65\% \\
(2) w/o Low-Res & 44.58 & 53.29 & 51.44 & -1.48\% \\
\midrule
\multicolumn{5}{c}{\textit{Task 2: General Vehicle Segmentation}} \\
\midrule
\textbf{Aerial-R1 (Full)} & \textbf{57.53} & 74.52 & \textbf{68.86} & - \\
(3) w/o Consistency & 56.40 & \textbf{74.83} & 67.46 & -1.13\% \\
(4) w/o Low-Res & 56.49 & 73.72 & 68.55 & -1.04\% \\
\bottomrule
\end{tabular}
}
\end{sc}
\end{small}
\end{center}
\vskip -0in
\end{table}

The results in Table \ref{tab:ablation} highlight the distinct roles of each component. In Task 1, removing the Consistency Reward (Row 1) causes a substantial performance drop, with mIoU decreasing by 1.65\% and Recall dropping by 1.54\%. This confirms that enforcing group consensus is critical for reasoning about correct targets in ambiguous contexts. Similarly, the impact of the Low-Res stream is profound across both tasks. Removing it (Row 2 and Row 4) consistently degrades performance across almost all metrics, leading to a 1.48\% mIoU drop in Task 1 and a 1.04\% mIoU drop in Task 2. This systematic degradation proves that the Low-Res stream does more than just explicitly incentivize refusal (anti-hallucination); it effectively acts as a rigorous data augmentation strategy within the group-relative advantage computation. By training on degraded views, the model is forced to extract a more robust and generalized feature representation to identify targets under adverse conditions. Ultimately, the full model achieves a superior balance, delivering the highest mIoU of 57.53\% in Task 2 by successfully integrating both consistency and explicitly incentivized uncertainty.

\subsection{Extension: Generalized Reasoning Evaluation}
\label{sec:reasoning}

\begin{figure}[htb]
\vskip -0in
\begin{center}
\centerline{\includegraphics[width=1\columnwidth]{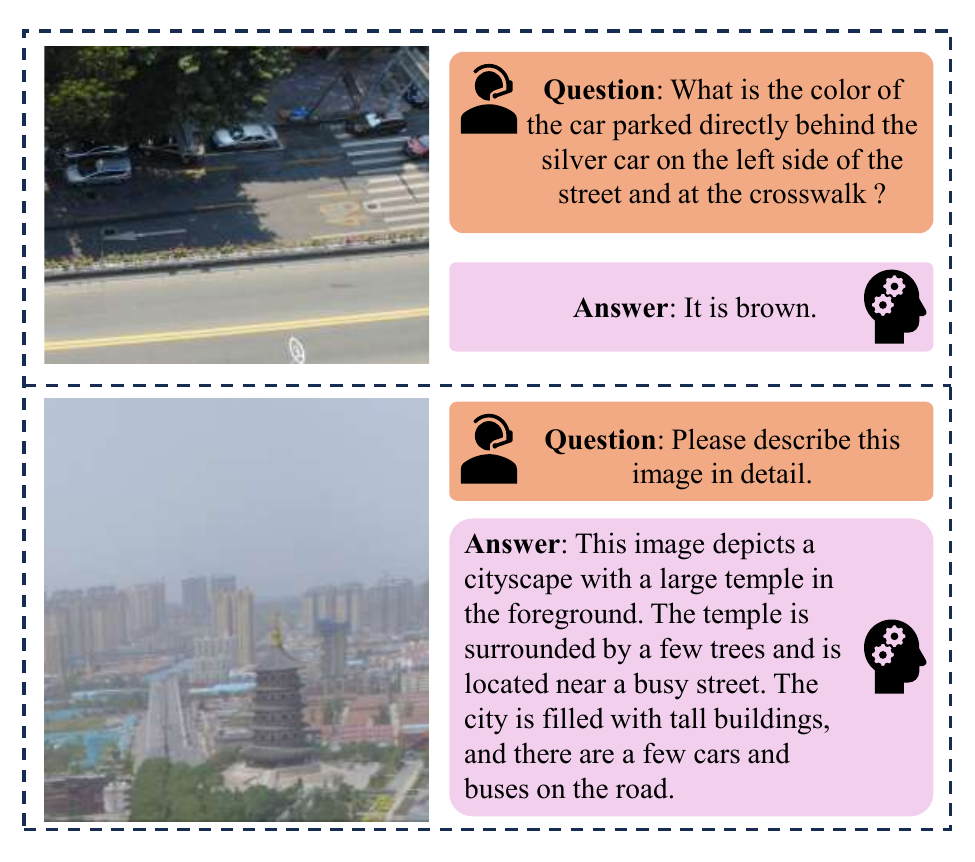}}
\caption{Qualitative examples of generalized reasoning. {(Top)} The model successfully resolves a complex multi-hop VQA query requiring sequential spatial grounding: identifying the \textit{silver car} at the \textit{crosswalk} on the \textit{left side}, and then reasoning about the object \textit{directly behind} it to retrieve its color attribute. {(Bottom)} A comprehensive scene description demonstrating the model's capacity to understand global context, accurately capturing key landmarks (the temple), surrounding environments, and spatial relationships. These examples collectively demonstrate that Aerial-R1 retains robust conversational and holistic descriptive capabilities alongside dense segmentation precision.}
\label{fig:vqa_reasoning}
\end{center}
\vskip -0.3in
\end{figure}

A major concern in aligning MLLMs for specific dense prediction tasks is the potential catastrophic forgetting of general conversational abilities \cite{kirkpatrick2017overcoming}. 
To verify that Aerial-R1 retains its reasoning capability, we evaluated the model on a free-form Visual Question Answering (VQA) benchmark and a holistic Scene Description task derived from VSAI-Ref. 
The VQA benchmark aggregates various reasoning tasks, including attribute identification and numerical reasoning, adapting the evaluation protocol from POPE \cite{li2023pope} and TallyQA \cite{acharya2019tallyqa}. 
To vividly demonstrate this capability, we provide a qualitative example in Figure \ref{fig:vqa_reasoning}, where the model successfully resolves a complex multi-hop spatial query.

\begin{table}[htb]
\caption{Generalized Reasoning Performance. We report the overall accuracy (\%) on the VSAI-Ref VQA benchmark, along with ROUGE-L and CIDEr scores (scaled by 100) for the Scene Description task. Best results are highlighted in bold.}
\label{tab:vqa_reasoning}
\centering 
\small
\begin{sc}
\resizebox{\linewidth}{!}{
\begin{tabular}{l|c|cc}
\toprule
\multirow{2}{*}{Method} & \multicolumn{1}{c|}{VQA} & \multicolumn{2}{c}{Scene Description} \\
& Acc. (\%) & ROUGE-L & CIDEr \\
\midrule
Sa2VA (SFT) \cite{yuan2025sa2va} & 69.83 & 36.40 & 37.88 \\
Sa2VA (SFT + Low-Res Aug.) & 69.48 & 32.41 & 31.71 \\
\textbf{Aerial-R1 (Ours)} & \textbf{71.73} & \textbf{37.83} & \textbf{53.99} \\
\bottomrule
\end{tabular}
}
\end{sc}
\vskip -0in
\end{table}

As presented in Table \ref{tab:vqa_reasoning}, Aerial-R1 demonstrates superior performance across all generalized reasoning metrics. On the VQA task, it achieves an accuracy of 71.73\%, outperforming the standard SFT baseline (+1.90\%). More impressively, in the complex Scene Description task, Aerial-R1 achieves a massive improvement in the CIDEr metric, jumping from 37.88 to 53.99. Crucially, the naive Sa2VA (SFT + Low-Res Aug.) baseline actively degrades conversational capabilities, causing a severe drop in both ROUGE-L and CIDEr scores. This further proves that our advantage-weighted extraction strategy not only refines pixel-level grounding but also enhances the model's ability to interpret and reason about visual content in a broader context, effectively mitigating catastrophic forgetting while boosting global contextual understanding.

\subsection{Training Dynamics \& Stability Analysis}
\label{sec:training_dynamics}

\begin{figure}[htb]
\vskip -0.1in
\begin{center}
\centerline{\includegraphics[width=0.9\columnwidth]{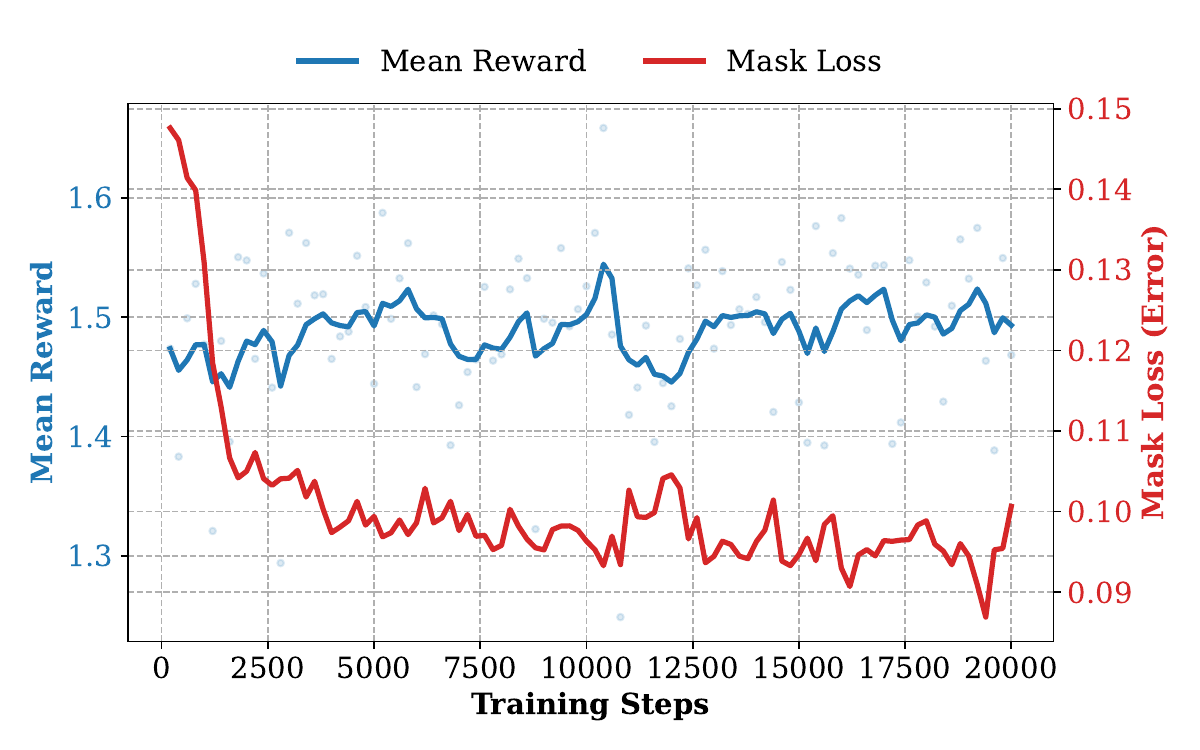}} 
\caption{Training Dynamics of Aerial-R1. The red line (right axis) tracks the Mask Loss, showing a clear downward trend indicative of effective, stable convergence under the AWR objective. The blue line (left axis) represents the Mean Reward, which rapidly plateaus ($>$1.4), demonstrating that the model quickly learns and maintains the safety boundary despite the discrete nature of the anti-hallucination penalty.}
\label{fig:training_dynamics}
\end{center}
\vskip -0.2in
\end{figure}

A common concern with Reinforcement Learning, particularly when applying policy optimization to non-autoregressive decoders, is training instability. To demonstrate the robust nature of our Advantage-Weighted Regression approach, we analyze the training dynamics in Figure \ref{fig:training_dynamics}. As shown in the figure, the mask loss exhibits a sharp decline in the initial phase (0-2k steps) and subsequently converges to a low error plateau of approximately $0.095$. The advantage-weighted objective successfully isolates the high-quality trajectories (where $A_i > 0$), allowing the model to minimize pixel-level errors smoothly without the divergence commonly seen in standard policy gradient methods.

Furthermore, the mean reward maintains a high plateau. While minor fluctuations and an early plateau are observed (an expected phenomenon due to the highly discrete and strict nature of our anti-hallucination mechanism, which switches rigidly between a $+1.0$ silence reward and a $-2.0$ penalty), the overall trajectory confirms that the policy rapidly discovers and settles into a stable equilibrium. It consistently enforces the learned safety boundary, choosing to refuse rather than hallucinate on non-existent objects.

\subsection{Limitations and Failure Case Analysis}
\label{sec:failure}

Despite achieving state-of-the-art performance, Aerial-R1 exhibits limitations in certain extreme scenarios, primarily manifesting in two failure modes. First, under severe occlusion, such as when a vehicle is more than $80\%$ obscured by trees or overpasses, the model tends to trigger its explicitly incentivized refusal mechanism and output an empty mask. While this leads to a false negative, it highlights a strict trade-off where the model chooses to be conservative rather than risking a wrong guess. Second, fine-grained attribute classification remains challenging. Distinguishing between visually similar vehicle categories (e.g., \textit{MPV} vs. \textit{SUV}) or dark colors (e.g., \textit{black} vs. \textit{dark blue}) is inherently difficult in low-resolution aerial views. In such cases, the model may occasionally fall back on language priors rather than visual features, leading to attribute mismatch errors.

\section{Conclusion}

In this work, we presented Aerial-R1, a pioneering framework bridging reasoning-centric Reinforcement Learning and trustworthy aerial visual perception. 
By formulating the optimization as a Hybrid-View Group Advantage-Weighted Regression, we overcome the mathematical intractability of applying token-level policy gradients to non-autoregressive mask decoders. This strategy successfully endows the model with the ability to reason about visual consistency across varying resolutions rather than merely memorizing semantic mappings. 
Our extensive experiments on the VSAI-Ref benchmark demonstrate that this approach not only achieves state-of-the-art precision on tiny objects but also establishes a robust, explicitly incentivized safety boundary, enabling the model to confidently refuse predictions when visual evidence is insufficient. 
We believe this paradigm shift from pure supervised fitting to advantage-weighted visual reasoning paves the way for the next generation of trustworthy and autonomous aerial agents capable of operating in complex, open-world environments.

\bibliographystyle{IEEEtran}
\bibliography{references}

\section*{Author Biography}
\authorbibliography[scale=0.09,overhang=0pt,wraplines=9,imagewidth=3cm,imagepos=l]{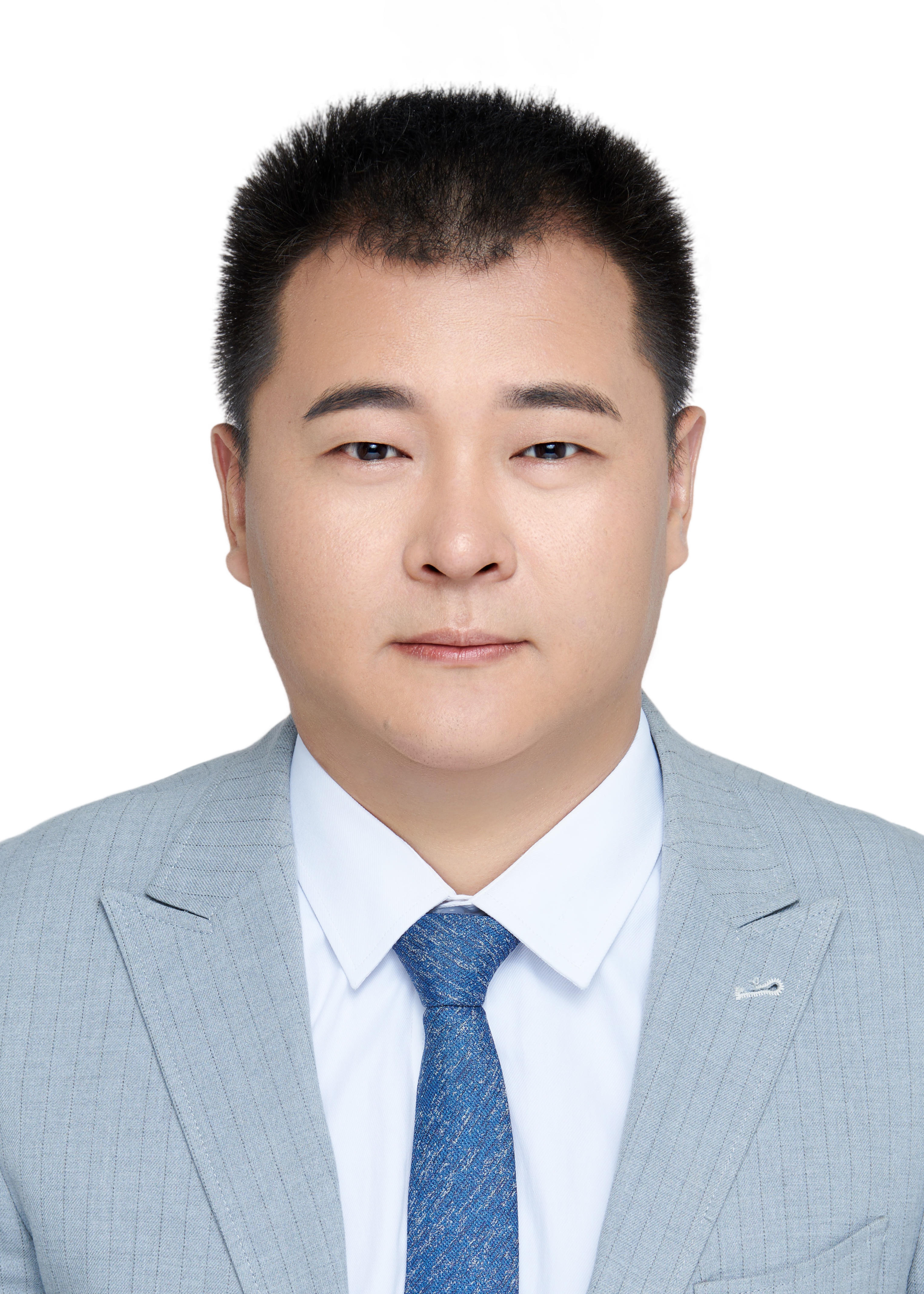}{Jisheng Dang \rm{
received the Ph.D. degree from Sun Yat-sen University, China, in 2025, advised by prof. Jianhuang Lai and prof. Huicheng Zheng.
He worked as a research fellow at the NExT++ laboratory of the NationalT University of Singapore  advised by prof. Tat-Seng Chua. He is now a tenured associate professor at the School of Information Science and Engineering, Lanzhou University. His research interests include multimodal learning, video understanding, and embodied  intelligence. 
He has published several papers as the first author in major journals and conferences including IEEE TIP/TNNLS/TITS/IJCAI/AAAI. He served as a reviewer at some major journals and conferences like IEEE TPAMI, ICML, NIPS, ICLR, IEEE TIP, CVPR, IJCAI, ACM MM, AAAI, IEEE TMM, IEEE TCSVT, ACM TOMM.}}\\

\authorbibliography[scale=0.12,overhang=0pt,wraplines=9,imagewidth=3cm,imagepos=l]{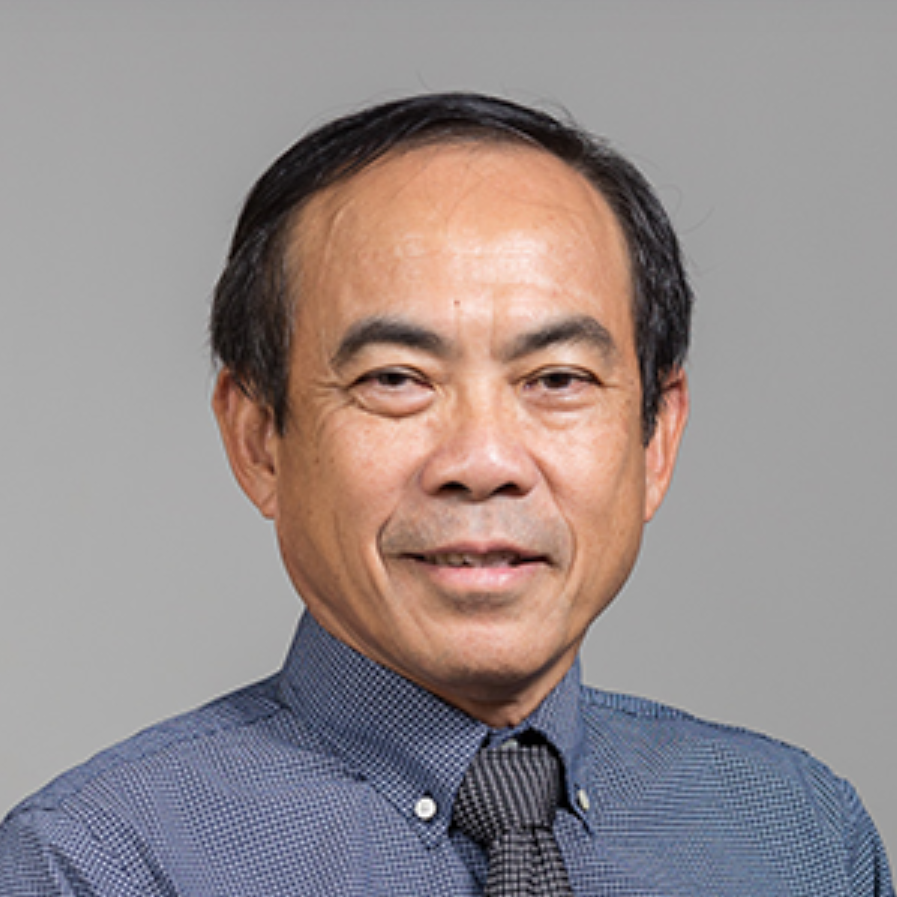}{Tat-Seng Chua \rm{received the Ph.D. degree from the University of Leeds, U.K. He is the KITHCT chair  professor with the School of Computing, National University of Singapore, where he was the acting and  founding dean of the School from 1998 to 2000.  He  is the co-director of NExT, a joint center between NUS and Tsinghua University, to develop technologies for live social media  search. He is the 2015 winner of the prestigious ACM SIGMM Award. He is the chair of Steering Committee of the ACM International Conference on Multimedia Retrieval (ICMR) and Multimedia Modeling (MMM) conference series.   He is also the general co-chair of ACM Multimedia 2005, ACM CIVR (now ACM ICMR) 2005, ACM SIGIR 2008, and ACM Web Science 2015.   He serves on the editorial boards of four international  journals. He is the co-founder of two technology startups in Singapore and a Fellow of the Singapore Academy of Sciences, with    108,708 citations on Google Scholar.}}\\

\end{document}